\documentclass[conference]{IEEEtran}
\IEEEoverridecommandlockouts

\usepackage[backend = bibtex,
			style = numeric,
			sorting = none,
			]{biblatex}
	
\usepackage{amsmath,amssymb,amsfonts}
\usepackage{graphicx}
\usepackage{textcomp}
\usepackage{xcolor}
\usepackage{url}
\usepackage{times}
\usepackage{epsfig}
\usepackage{subfigure} 
\usepackage{algorithm}
\usepackage{caption}
\usepackage[algo2e]{algorithm2e}
\usepackage[]{algpseudocode}
\usepackage{algorithm2e}
\usepackage{setspace}
\usepackage{multirow}
\usepackage{lettrine}

{\footnotesize
\bibliography{egbib_15.bib, caleb-zotero.bib}}

\def\BibTeX{{\rm B\kern-.05em{\sc i\kern-.025em b}\kern-.08em
    T\kern-.1667em\lower.7ex\hbox{E}\kern-.125emX}}

\begin{document}

\title{ On the fly Deep Neural Network Optimization Control for Low-Power Computer Vision
}

\author{Ishmeet Kaur, Adwaita Janardhan Jadhav
}

\maketitle

\begin{abstract}

Processing visual data on mobile devices has many applications, e.g., emergency response and tracking. State-of-the-art computer vision techniques rely on large Deep Neural Networks (DNNs) that are usually too power-hungry to be deployed on resource-constrained edge devices. Many techniques improve the efficiency of DNNs by using sparsity or quantization. However, the accuracy and efficiency of these techniques cannot be adapted for diverse edge applications with different hardware constraints and accuracy requirements.
This paper presents a novel technique to allow DNNs to adapt their accuracy and energy consumption during run-time, without the need for any re-training. Our technique called AdaptiveActivation introduces a hyper-parameter that controls the output range of the DNNs' activation function to dynamically adjust the sparsity and precision in the DNN. AdaptiveActivation can be applied to any existing pre-trained DNN to improve their deployability in diverse edge environments. We conduct experiments on popular edge devices and show that the accuracy is within 1.5\% of the baseline. We also show that our approach requires 10\%--38\% less memory than the baseline techniques leading to more accuracy-efficiency tradeoff options. 

\end{abstract}


\section{Introduction}

Deep Neural Networks (DNNs) are important tools in computing. They are heavily used in fields like computer vision~\cite{covid-1, mottaghi_role_2014}.
DNNs are not just simple neural networks with a couple hundred neurons, they 
are made of many layers with a large spread of connections between layers. This gives DNNs a tremendous range of variability that can be fine-tuned for accurate inference through training. 
Unfortunately, DNNs are also computation-heavy and energy-expensive as a result. ResNet~\cite{resnet}, a popular DNN used in computer vision, needs to perform billions of operations and general gigabytes of memory to perform image classification on a single image~\cite{han2016}. These many computations and memory operations require significant compute resources and lead to high energy costs~\cite{goel2020survey}.

\begin{figure}[t]
\begin{center}
\centering
\subfigure[]{\includegraphics[width = 0.49\textwidth]{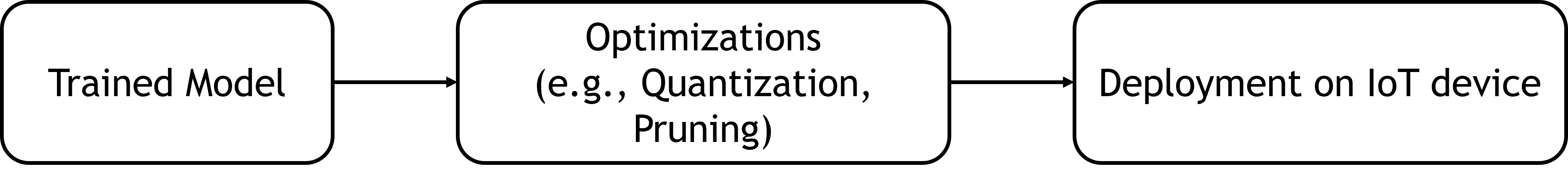}
}\\
\subfigure[]{\includegraphics[width = 0.49\textwidth]{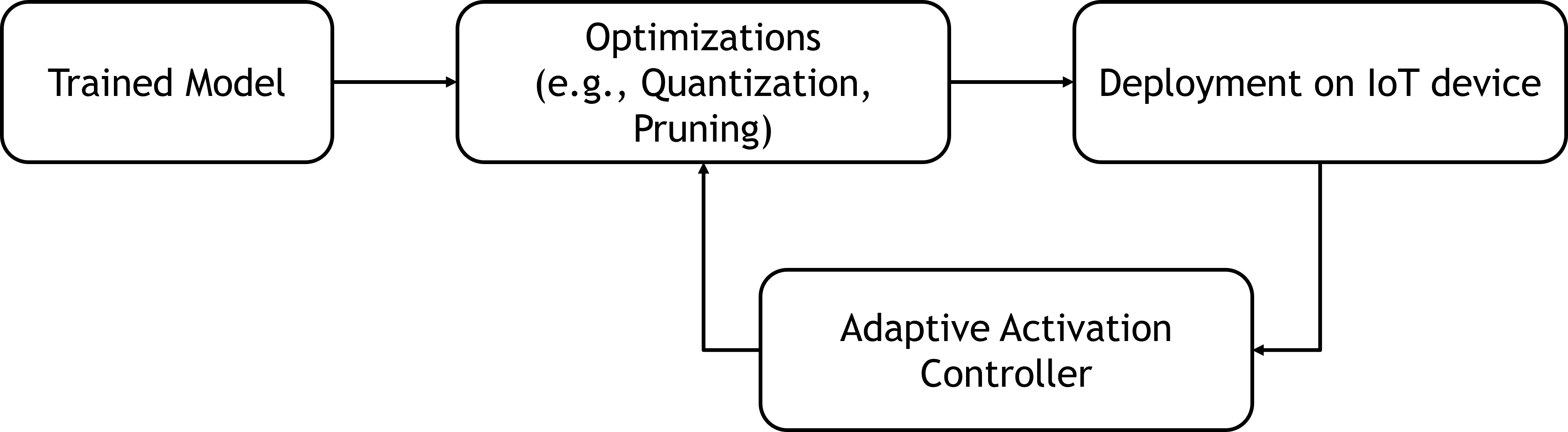}
}
\end{center}
   \caption{(a)~Existing low-power computer vision techniques apply optimizations before deployment, leading to a static accuracy-efficiency tradeoff. (b)~ The proposed method adds an Adaptive Activation Controller to adjust the DNNs' hyperparameters to change the accuracy-efficiency tradeoff dynamically during run-time.}
\label{fig:concurrent_vision_processing1}

\end{figure}

The excessive energy costs present a problem for DNNs --- it is difficult to deploy them in low-power embedded IoT environments. Most IoT devices are often constrained by battery power~\cite{anup} and have limited computing resources. To address this issue, many techniques have been developed to improve the efficiency of DNNs at the expense of accuracy~\cite{Han2015}. However, in most of the existing low-power computer vision techniques, the compromise between accuracy and efficiency cannot be tuned (depicted in Fig.~\ref{fig:concurrent_vision_processing1}(a)). This leads to DNNs that are either too inaccurate because of the accuracy losses, or DNNs that are too power-hungry because of the lack of optimizations~\cite{tnn}. Moreover, in real-world DNN applications, the ambient temperature varies widely based on the weather and the heat dissipation~\cite{L_Liu, 9680182}. As a result, the power and thermal budgets available for processing change dynamically~\cite{goel2020survey}. This also creates a need for dynamically tunable DNN workloads (depicted in Fig.~\ref{fig:concurrent_vision_processing1}(b)). 

In this work, we present a method to vary the accuracy-efficiency tradeoff configurations on the fly. Our technique, called AdaptiveActivation, can be applied to any existing DNN without any retraining. By using the available power budget as an input, AdaptiveActivation transforms the activation functions' output to increase or decrease the sparsity or precision in the DNN. As noted in prior research, higher sparsity or more precision in the DNN increases efficiency and decreases accuracy, and vice versa. Our method measures the DNN layers' sensitivity to sparsity and precision and uses that information to ensure that the accuracy losses are minimal. 


To evaluate this approach, our experiments compare the image classification accuracy and efficiency of the different techniques. This paper uses a computer vision dataset that contains varying input resolutions to show that this method can vary the accuracy-efficiency tradeoff for different workload types. We also deploy on hardware and use the Intel Power Calculator~\cite{Intel} to compare the energy consumption and latency. Using the proposed approach, we observe that our method reduces the memory of existing DNNs, such as MobileNet and ResNet-18, by upto $\sim$38\% without significant accuracy losses and the need for any re-training.

The rest of the paper is organized as follows: In Section~\ref{sec:related}, we highlight relevant existing work. In Section~\ref{sec:aa}, we present details about our technique. Section~\ref{sec:expt} contains information about our experimental setup and compares the results of our technique with the existing state-of-the-art. Finally, Section~\ref{sec:conc} concludes this paper.

\section{Related Work}
\label{sec:related}

\subsection{Deep Neural Networks for Computer Vision}
\label{sub:dnn}
Deep neural networks (DNNs) are a class of machine learning algorithms that can achieve high accuracy on many computer vision tasks~\cite{thiruvathukal_low-power_nodate, Howard_2019_ICCV, vgg}. DNNs use gradient descent to train models with millions or even billions of parameters with large amounts of training data. This is the main reason why DNNs are the most accurate computer vision techniques. DNNs generally contain several layers. The different layers are connected with activation functions. These activation functions provide non-linearity in the DNN helping the DNN converge better. Some commonly used activation functions are Sigmoid~\cite{NARAYAN199769} and ReLU~\cite{agarap2019deep}. In this paper, we propose a transformation to the existing ReLU activation function to enable dynamically changing accuracy vs. efficiency tradeoffs for edge applications.


\subsection{Low Power Computer Vision Techniques}

Goel et al.~\cite{goel2020survey} survey low-power DNNs and describe the benefits of reducing memory and operations for low-power applications. There are many DNN optimization techniques that increase DNN efficiency~\cite{thiruvathukal_low-power_nodate, 9502472, goel2020survey}. For example, MobileNet~\cite{Howard_2019_ICCV} is a DNN micro-architecture that uses bottleneck layers to reduce the number of parameters. Quantization and pruning~\cite{Han2015} are other commonly used techniques used to reduce the memory requirement of DNNs. NN quantization reduces the memory requirement and DNN pruning reduces the DNN operations~\cite{lpirc}. However, in such existing efficient DNN inference techniques, the compromise between accuracy and efficiency is fixed and cannot be tuned.  
We propose a method that can select the activation sparsity and quantization to vary the accuracy-efficiency tradeoff on-the-fly.


 

\subsection{Adaptive DNN Workloads}

Because of the importance of having adaptive workloads for different application scenarios and hardware requirements, a significant amount of research focuses on building on re-configurable DNN architectures. In these techniques, the DNNs use different layers and connections based on the input~\cite{gater, todaes}. GaterNet~\cite{gater} is a technique that uses a gating layer to deactivate certain DNN connections to reduce computation costs. Other techniques~\cite{todaes, tnn} use different connections in the DNN based on the input to decrease the number of arithmetic operations. These techniques add new components or layers to the DNN and often require expensive training and/or re-training schemes. Our proposed method requires no retraining and still can be adapted for various accuracy vs. efficiency tradeoff configurations on the fly.



\subsection{Our Contributions}

(1)~This is the first method to tune the accuracy-efficiency tradeoff by transforming the output of the activation function to control sparsity and quantization. Our method can be applied to any existing pre-trained DNN without the need for any re-training.
(2)~We develop methods to select the optimal sparsity and quantization levels for each DNN layer based on their sensitivity.
(3)~Experiments show that our method outperforms the baseline techniques in terms of memory, number of operations, and energy.

\section{AdaptiveActivation: Dynamic Control of DNNs}
\label{sec:aa}
\begin{figure}[t]
\begin{center}
\includegraphics[width=0.95\linewidth]{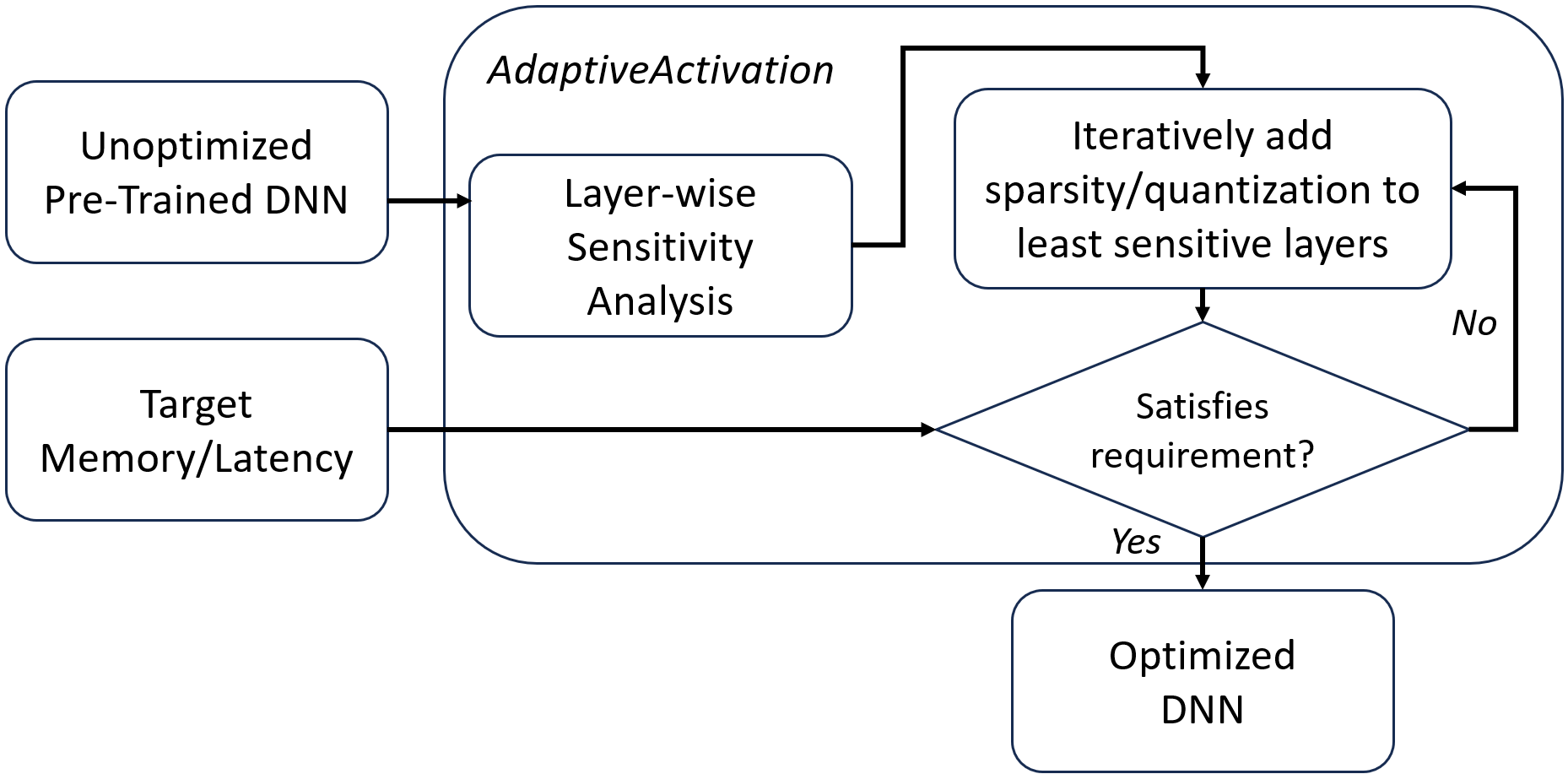}
\end{center}
\caption{Proposed AdaptiveActivation method that can control the accuracy vs. efficiency tradeoff on the fly.}
\label{fig:overall_flow}
\end{figure}

Our proposed method, called AdaptiveActivation, transforms the output of the DNN's activation function to change the sparsity and quantization levels during inference. Fig.~\ref{fig:overall_flow} outlines the working of AdaptiveActivation. The inputs are a pre-trained DNN and the target memory/latency requirements of the activation. AdaptiveActivation first calculates each layer's sensitivity to varying sparsity and quantization. Using this information, AdaptiveActivation iteratively adds sparsity and reduces the precision (quantization level) of the least sensitive layers. The least sensitive layers are chosen because they can be optimized with minimal impact on the DNN accuracy. In the following subsections, we provide more details about each component used in AdaptiveActivation.

\begin{figure*}[t]
\begin{center}
\centering
\subfigure[]{\includegraphics[width = 0.35\textwidth]{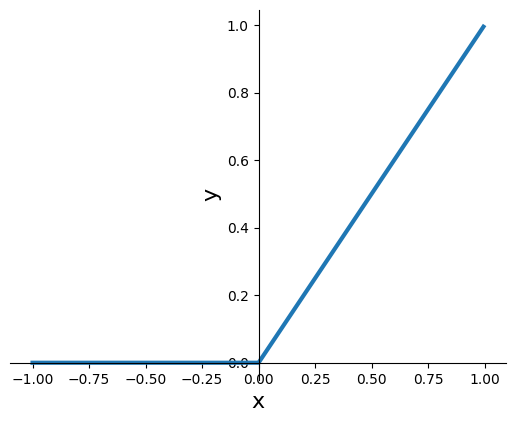}
}
\subfigure[]{\includegraphics[width = 0.35\textwidth]{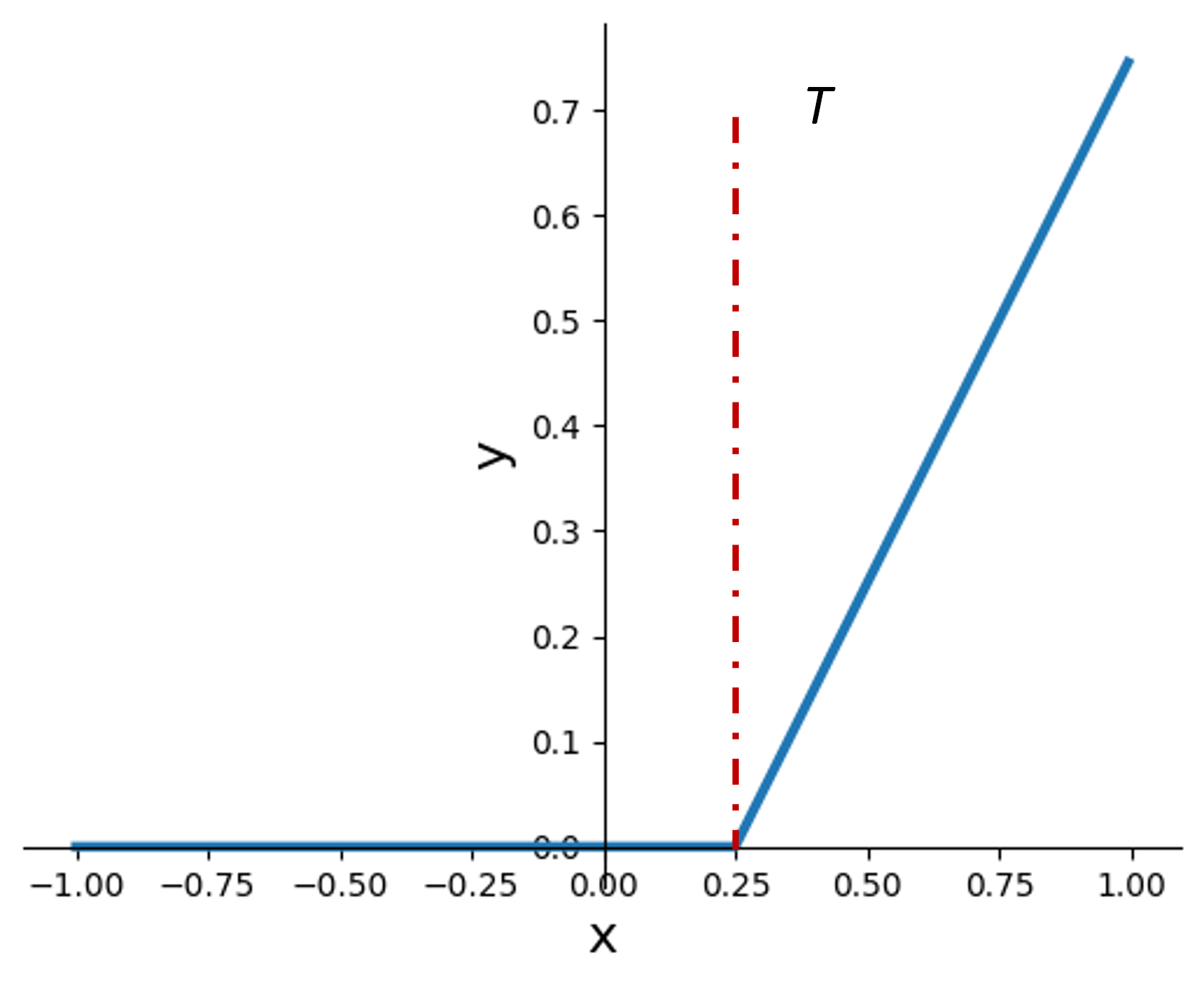}
}
\end{center}
   \caption{(a)~In the existing ReLU activation function, all values greater than zero are retained, and values smaller than zero are converted to zero. (b)~In our proposed AdaptiveActivation ReLU (AA-ReLU) activation the activation function's cutoff $T$ is adjustable for each DNN layer. This allows us to increase or decrease the sparsity in the DNN.}
\label{fig:concurrent_vision_processing}

\end{figure*}

\subsection{Controlling Activation Sparsity and Precision}

In this section, we highlight our methods to control the activation sparsity and precision in DNNs.

\begin{figure}[b!]
\begin{center}
\includegraphics[width=0.8\linewidth]{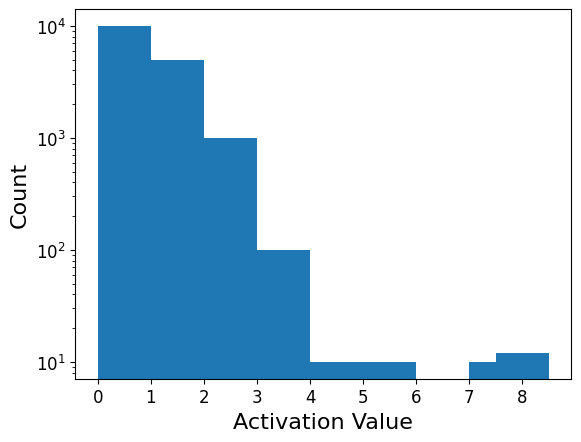}
\end{center}
\caption{Activation output histogram for one layer of the pre-trained VGG-16 DNN averaged over 128 inputs. A significant portion of the activation values are zero.}
\label{fig:hist}
\end{figure}

\subsubsection{AdaptiveActivation ReLU}

In this section, we describe our method to control the sparsity in DNN activations. We specifically focus on methods that require no no-retraining because they can be used for on-the-fly adjustments to inference performance. 

Fig.~\ref{fig:concurrent_vision_processing}(a) and Eqn.~(\ref{eqn:relu}) depict the existing ReLU activation function. In the existing activation function, only the input values ($x$) below zero are converted to zero in the output. In order to control the sparsity of the existing ReLU, we add a new hyper-parameter $T$. AdaptiveActivation ReLU (AA-ReLU) is depicted in Fig.~\ref{fig:concurrent_vision_processing}(b) and Eqn.~(\ref{eqn:aarelu}). By adjusting the value of $T$, we can control the sparsity in the DNN.

\begin{equation}
    y =
    \left\{\begin{matrix}
        x & \text{if } x > 0\\
        0 & \text{otherwise}
    \end{matrix}\right.
    \label{eqn:relu}
\end{equation}

\begin{equation}
    y =
    \left\{\begin{matrix}
        x-T & \text{if } x > T\\
        0 & \text{otherwise}
    \end{matrix}\right.
    \label{eqn:aarelu}
\end{equation}

\begin{figure}[b!]
\begin{center}
\includegraphics[width=0.9\linewidth]{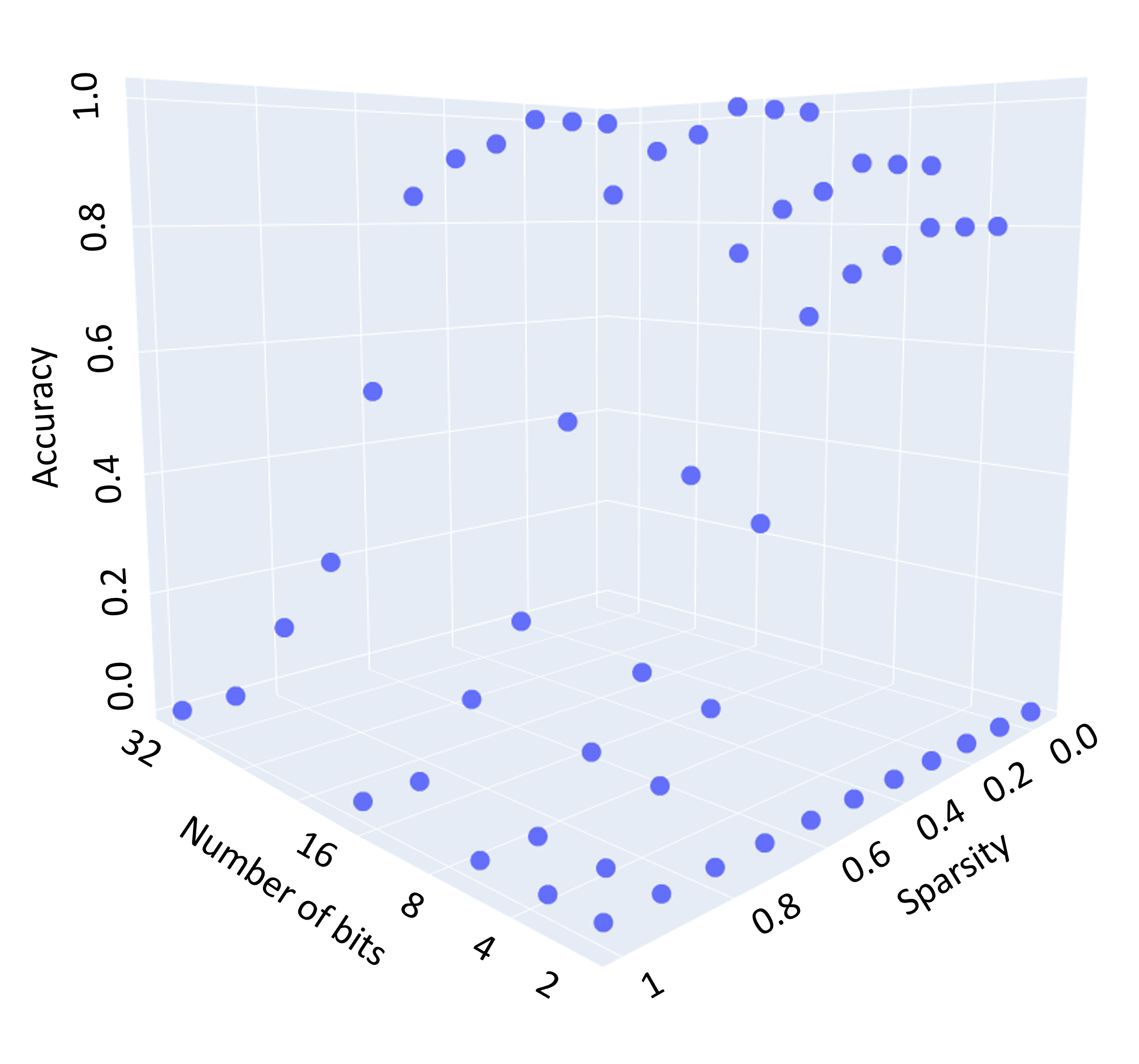}
\end{center}
\caption{Activation sensitivity analysis on one layer of the VGG-16 DNN. The sparsity axis measures the additional sparsity added to the DNN activation (0.0 means no additional sparsity and 1.0 means all values are rounded to 0). The number of bits axis represents the precision of the activation outputs (the default value is 32-bits). The accuracy axis measures the relative accuracy of the overall DNN when this layer is set to a specific precision and sparsity.}
\label{fig:plotly}
\end{figure}

\subsubsection{Activation Precision Control}

The default quantization format used in DNN libraries such as PyTorch is 32-bit floating point. Howver, many other quantization formats can be used for inference. Some quantization formats are: 16-bit floating point, 8-bit floating point, 4-bit integer, and 2-bit integer. Using fewer bits reduces the memory requirement, and also speeds up the arithmetic operations. However, using fewer bits reduces the accuracy of the DNN.

In this work, we use different quantization levels for different layers based on the impact on the accuracy. Cast operators are inserted at the output of the activations to convert values from one quantization level to another.

\subsection{Layer-wise Sensitivity Analysis}

Upon analyzing the distributions of activations in the layers in DNN, it is evident that a significant portion of the values are rounded to zero. This is because of the use of the ReLU activation function (depicted in Fig.~\ref{fig:concurrent_vision_processing}(a)). Fig.~\ref{fig:hist} depicts a histogram analysis for the VGG-16 DNN. 
This provides evidence that DNN's activation functions can handle sparse feature maps. Our work uses the AA-ReLU function (depicted in Fig.~\ref{fig:concurrent_vision_processing}(b)) to further increase the DNN sparsity to improve efficiency, without significant accuracy losses.

However, different layers have different distributions and thus have varied sensitivities to sparsity and precision. 
Fig.~\ref{fig:plotly} depicts our
activation sensitivity analysis for the same layer in the VGG-16 DNN. In the activation sensitivity analysis, for each layer independently, we vary the sparsity and precision (using AA-ReLU) and measure the impact on the overall DNN accuracy on the testing dataset. When performing the sensitivity analysis on a particular layer, all other layers are frozen. This allows us to optimally set the sparsity and precision for each DNN layer using the algorithm described in the next subsection to ensure the best tradeoff between accuracy and efficiency.

\begin{figure}[b!]
\begin{center}
\includegraphics[width=0.75\linewidth]{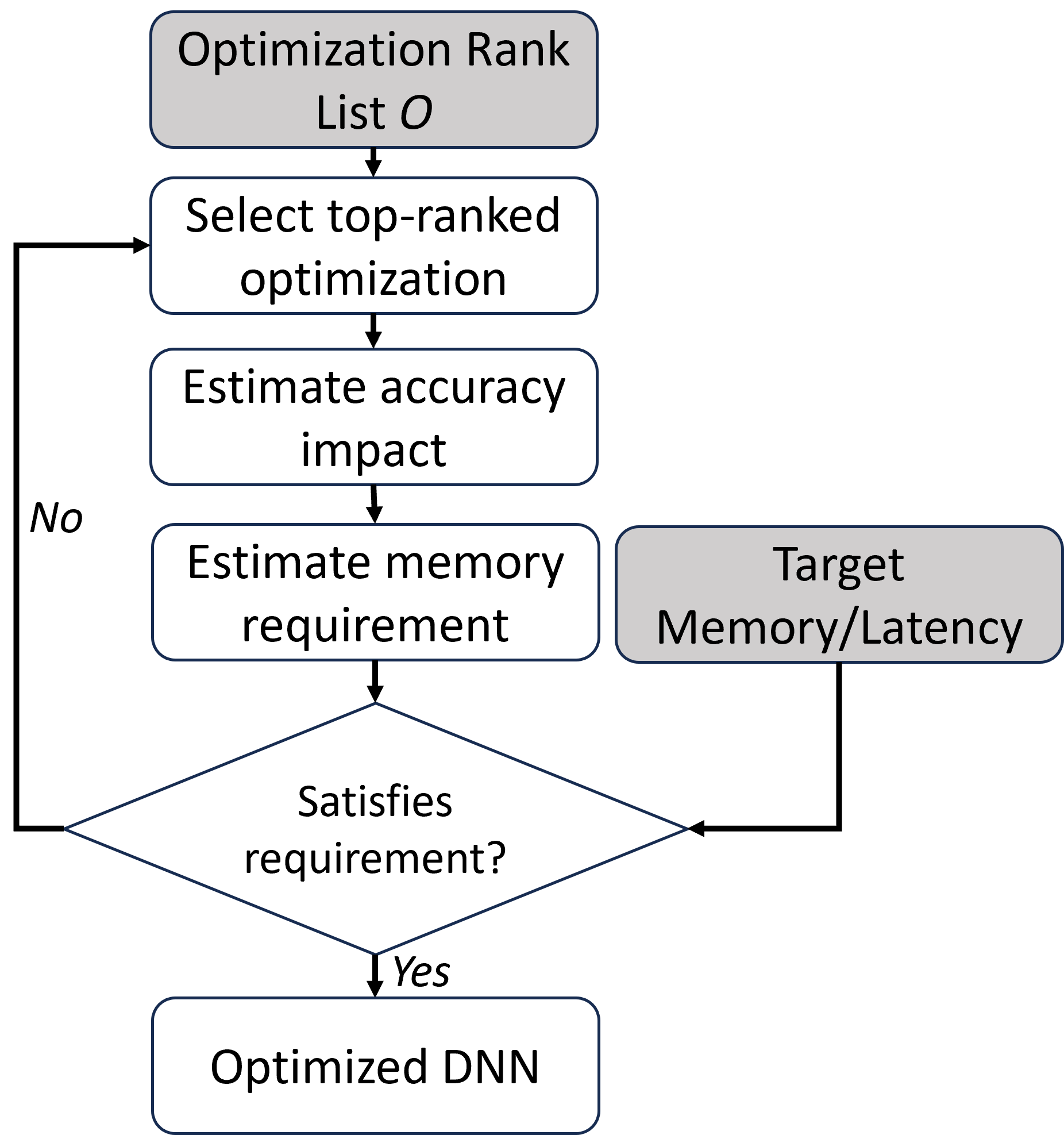}
\end{center}
\caption{Algorithm to greedily select optimizations to meet the target memory/latency requirement. The grey boxes depict the input to the algorithm. The optimized DNN is the output.}
\label{fig:alg}
\end{figure}

\subsection{Optimally Selecting Per-Layer Sparsity and Precision for a Given Memory Budget}



The task is to select the sparsity layers and precision levels for each DNN layer in a way that minimizes the accuracy loss and meets the target performance requirements. We propose a greedy iterative method to update the hyper-parameters (i.e., the sparsity and precision) one layer at a time. The greedy selection selects $S$ pre-defined sparsity levels and $Q$ pre-defined precision levels (as depicted in the sensitivity analysis in Fig.~\ref{fig:plotly}). For each layer, the $S \times Q$ hyper-parameter combinations are evaluated for their impact on memory and accuracy. Across the model, a rank-list, $O$ of most suitable optimizations is then created, where the top-ranked optimization has the least impact on accuracy and the most memory improvement. Our greedy algorithm iteratively selects the optimizations from $O$ with the highest score until the target memory requirement is reached. The algorithm is depicted in Fig.~\ref{fig:alg}.

\section{Experiments and Results}
\label{sec:expt}

This section shows that a PyTorch implementation of the AdaptiveActivation technique improves efficiency when compared to the baseline without sacrificing classification accuracy. Through other experiments, we also show that our method offers multiple accuracy-efficiency tradeoff options. We will open-source our code with the final version of the paper.

\vspace{-0.1in}
\subsection{Datasets Used}

We use the ImageNet dataset~\cite{5206848} in our experiment. This is a commonly used dataset dataset that has varied input data (greyscale and color) of different-sized inputs that showcases the generality of our technique.


\subsection{Evaluation Metrics} 

The image classification accuracy is used as the primary metric. The classification accuracy is the number of images classified correctly from the test set divided by the total number of images. To compare the efficiency, we use the memory requirement (model size + activation memory size). The memory requirement is calculated using the number of parameters and activations multiplied by the number of bytes used (i.e., the precision). We use the Intel Power Calculator~\cite{Intel} to estimate the energy requirements and latency of the PyTorch programs running on Intel CPUs.

\begin{table}[t!]
\begin{tabular}{|c|l|rr|rr|}
\hline
\multirow{2}{*}{\textbf{Dataset}} & \multicolumn{1}{c|}{\multirow{2}{*}{\textbf{DNN}}} & \multicolumn{2}{c|}{\textbf{Baseline}} & \multicolumn{2}{c|}{\textbf{AA}} \\ \cline{3-6} 
 & \multicolumn{1}{c|}{} & \multicolumn{1}{c|}{\textbf{Acc.}} & \multicolumn{1}{c|}{\textbf{Mem.}} & \multicolumn{1}{c|}{\textbf{Acc.}} & \multicolumn{1}{c|}{\textbf{Mem.}} \\ \hline
\multirow{3}{*}{ImageNet} & ResNet-18 & \multicolumn{1}{r|}{72.4\%} & 68 & \multicolumn{1}{r|}{71.0\%} & 55 \\ \cline{2-6} 
 & MobileNet & \multicolumn{1}{r|}{72.0\%} & 54  & \multicolumn{1}{r|}{70.1\%} &  33 \\ \cline{2-6} 
 & VGG-16 & \multicolumn{1}{r|}{70.5\%} &  570 & \multicolumn{1}{r|}{69.8\%} & 510 \\ \hline
\end{tabular}
\caption{Comparison of accuracy and memory (in MB) of three popular DNN architectures with and without AdaptiveActivation (AA).}
\label{tab:one}
\end{table}

\subsubsection{Comparison with baselines}

When comparing with baseline techniques, we consider 3 popular DNN architectures: ResNet-18~\cite{resnet}, VGG-16~\cite{vgg}, and MobileNet~\cite{Howard_2019_ICCV}. These architectures are representative of most convolutional DNN architectures used today. We hypothesize that this technique can be applied to Transformer-based architectures, but is not included in this paper and will be considered in future work.

Table~\ref{tab:one} compares the image classification accuracy and memory requirements of the three baseline DNNs with and without AdaptiveActivation. Since AdaptiveActivation allows for many different energy-efficiency tradeoffs, we present one tradeoff in Table~\ref{tab:one}.
Our method requires 38\% ($1- \frac{33}{54}$) and 10.5\% ($1- \frac{510}{570}$)  less memory than the baseline MobileNet and VGG-16 implementations in PyTorch. The memory improvements in MobileNet are greater because MobileNet creates more activations due to the use of Depthwise Separable Convolutions. Across all three DNNs, the accuracy losses are small, i.e., 1\%-1.5\% degradation. This may be an acceptable tradeoff in many applications, especially because our technique does not require any re-training.

\subsubsection{Tunable Accuracy vs. Efficiency Tradeoff}

\begin{figure}[t!]
\begin{center}
\includegraphics[width=0.8\linewidth]{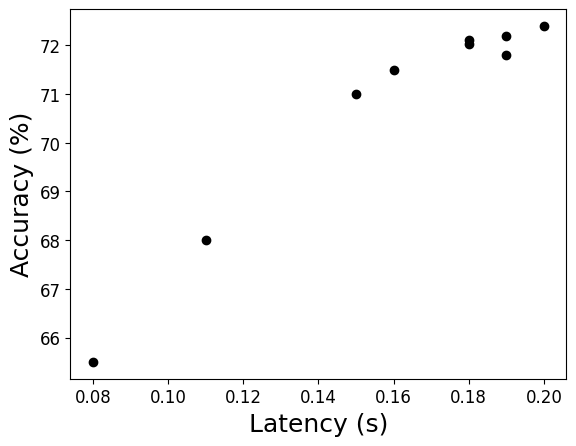}
\end{center}
\caption{Varying accuracy vs. efficiency (measured as latency) configurations possible with AdaptiveActivation when applied to the ResNet-18 DNN. The baseline ResNet-18 DNN requires 0.2 seconds to process an image (of size 224$\times$224 pixels) and archives an accuracy on 72.4\%.}
\label{fig:final}
\end{figure}

We deploy the PyTorch implementations on an Intel Core i7 CPU, with an average power of 28 W to compare the energy consumption and latency. We use the Intel Power Calculator tool to obtain the results. Fig.~\ref{fig:final} shows the various accuracy vs. efficiency (measured as latency) options made possible with AdaptiveActivation on the ResNet-18 DNN. Please note that all these tradeoff configurations are possible without the need for any retraining.

\section{Conclusions}
\label{sec:conc}

Through this paper, we solve an important problem in edge-based systems, i.e., most existing DNNs are too power-hungry to be deployed on most edge devices. Although there have been many techniques that optimize DNNs for low-power environments, they offer static optimizations, where the accuracy vs. efficiency tradeoff cannot be tuned. Such static optimizations lead to scenarios where the solution is accurate but too power-hungry, or are well-optimized but not accurate enough to be deployed in different application settings. To overcome this problem, we propose AdaptiveActivation. This is a technique that transforms the output of the DNN's activation function to increase the sparsity and/or reduce the precision to change the accuracy vs efficiency tradeoff. We also present an algorithm that can control the amount of sparsity/quantization added to the network to change the tradeoff on the fly without the need for any retraining. In our experiments, we consider three popular DNN architectures and showcase the ability of our technique to reduce the memory requirements and latency, with only small losses in accuracy. In future work, we will look to extend this research to apply to Transformer-based architectures as well.

\def\bibfont{\footnotesize}
\printbibliography

\end{document}